\documentclass[conference]{IEEEtran}

\ifCLASSINFOpdf
\else
\fi
\usepackage{algorithmic}
\usepackage{graphicx}
\usepackage{textcomp}
\usepackage{xcolor}
\usepackage{multirow}
\usepackage{tabularx,ragged2e}
\usepackage{tabulary}
\usepackage{adjustbox}
\usepackage{arydshln}
\usepackage[official]{eurosym}
\usepackage{cleveref}
\usepackage[flushleft]{threeparttable}
\usepackage{xurl}
\usepackage{lipsum}
\usepackage{graphicx}

\begin{document}
%
\title{An Interdisciplinary Approach for the Automated Detection and Visualization of Media Bias in News Articles}

\author{\IEEEauthorblockN{Timo Spinde}
\IEEEauthorblockA{School of Electrical,\\
Information and Media Engineering\\
University of Wuppertal\\
Email: timo.spinde@uni-wuppertal.de}}


%


\maketitle

\begin{abstract}
 Media coverage has a substantial effect on the public perception of events. Nevertheless, media outlets are often biased. One way to bias news articles is by altering the word choice. The automatic identification of bias by word choice is challenging, primarily due to the lack of gold-standard data sets and high context dependencies. In this research project, I aim to devise data sets and methods to identify media bias. To achieve this, I plan to research methods using natural language processing and deep learning while employing models and using analysis concepts from psychology and linguistics. The first results indicate the effectiveness of an interdisciplinary research approach. My vision is to devise a system that helps news readers become aware of media coverage differences caused by bias. So far, my best performing BERT-based model is pre-trained on a larger corpus consisting of distant labels, indicating that distant supervision has the potential to become a solution for the difficult task of bias detection.  
\end{abstract}


%
\IEEEpeerreviewmaketitle

\section{Introduction}\label{Intro}
News articles serve as a highly relevant source of information on current topics, and salient political issues \cite{dallmann2015a}\footnote{Please note that this proposal partially uses contents from my already published papers, mainly from \cite{Spinde2021, spinde2021mbic, spinde_how_2021}}. A varying word choice in any news content may have a significant effect on the public and individual perception of societal issues, especially since regular news consumers are mostly not fully aware of the degree and scope of bias \cite{spinde2020b, SpindeOmission}. As shown in existing research~\cite{park2009newscube, baumer2015a}, detecting and highlighting media bias might be relevant for media analysis and to mitigate the effects of biased reports on readers. Also, the detection of media bias can assist journalists and publishers in their work \cite{Spinde2021}. 

Bias, in general, is a highly sensitive topic, and some forms of bias especially rely on other factors than the content itself, such as a different perception of any text related to the individual background of a reader\cite{Spinde2021Embeddings}. As the concept and its perception are very interdisciplinary, its detection is a challenging task \cite{recasens2013a}. To date, only a few research projects, mostly based on linguistic features, focus on the detection and aggregation of bias \cite{lim2020annotating, hube2018detecting, recasens2013a}. Even though bias embodies a complex structure, contributions \cite{hube2019neural, chen_analyzing_2020} often neglect annotator background and use crowdsourcing to collect annotations. Therefore, existing data sets exhibit low annotator agreement and inferior quality.

The same low agreement in bias perception has also been researched from a psychological perspective: many individual factors affect the perception of bias, such as topic knowledge, political ideology, or simply age and education \cite{Spinde2021}. Phenomena like the Hostile Media Effect (HME, i.e., describing the tendency to perceive media coverage of an issue as biased against one's views \cite{gunther_assimilation_2009}) might also play a role, making it hard to objectively determine whether and how an article or clip is biased.

Various definitions and methods were used to measure media bias throughout the different studies on media bias perception and identification. Still, there exists a major lack of agreement on how study participants or readers react towards bias depending on how they were asked. Most existing studies focus only on specific aspects, for example, the already mentioned HME \cite{kim_partisans_2015, kim_role_2016, lee_liberal_2005}. Some studies asked questions related to particular articles \cite{gunther_broad_2006}, while others chose a more general approach \cite{glynn_how_2014}. Some ask about bias directly (e.g., "Regarding the web page that you viewed, would you say the portrayal of the presidential candidates was strictly neutral or biased in favor of one side or the other?" \cite{houston_influence_2011}), and some indirectly \cite{spinde2020b, spinde2020a, Spinde2020INRA, spinde_towards_2021}. Some researchers tried experiments [14], while others use surveys \cite{glynn_how_2014}. 

Overall, the psychological contributions focus more on the concept itself but lack scalability. The computer scientific approaches based on linguistics features are able to analyze large amounts of data but lack relevancy. Therefore, in this work, I will try to tackle the issue from an interdisciplinary perspective and build a system to identify media bias automatically. 

\section{Research Question, Tasks, and
Contributions}\label{sec:tasks}

My study holds both theoretical and practical significance. First, I will summarize the most influential work on media bias, to conclude a precise and suitable definition of a concept that both everyday readers as well as scientists from different areas can relate to and work with. Second, I will present various data sets measuring the concept, of which two are already published. Recently, I have published BABE (\textbf{B}ias \textbf{A}nnotations \textbf{B}y \textbf{E}xperts), a data set of media bias annotations, which is built on top of the previous MBIC (\textbf{M}edia \textbf{B}ias Annotation Data Set \textbf{I}ncluding Annotator \textbf{C}haracteristics) \cite{spinde2021mbic}. MBIC offers a balanced content selection, annotations on a word and sentence level, and is with 1,700 annotated sentences one of the largest data sets available in the domain. BABE improves MBIC, and other data sets, in two aspects. On the one hand, annotations are performed by trained experts and in a larger number. On the other hand, the corpus size is expanded considerably with additional 2,000 sentences.  The resulting labels are of higher quality and capture media bias better than labels gathered via crowdsourcing. In sum, BABE consists of 3,700 sentences with gold-standard expert annotations on the word and sentence level. In future work, I will extend the data set by using a gamified approach and publishing a third and final data set. 

To create these data sets and evaluate the final results, knowledge about the perception of media bias is crucial. Therefore, I have already analyzed which questions can reliably measure how a reader perceived bias and build a high-quality questionnaire for media bias assessment \cite{spinde2021B}. Additionally, I am researching how bias visualizations can both indicate bias for readers and teach them in a way that they become able to identify the concept on their own.

Lastly, I will implement and compare multiple models to identify bias automatically. To date, distant supervision appears to return the best results in my previous work. Still, other models might also improve performance, which is why a careful evaluation and comparison of available models and approaches are also part of my work. 

To focus on the system development as a whole, I define the following research question for my Ph.D. research: \textit{How can an automated approach identify media bias in English news articles?} For the classification presented in my work, I focus on sentence-level bias detection, which is the current standard in related work (\cref{sec:relatedwork}). To address my research question, I derive the following research tasks:

\begin{enumerate}
\item Review the state-of-the-art in identifying media bias (interdisciplinary) and identify the strengths and weaknesses of manual and automated methods used to identify media bias using computer science methods.
\item Research NLP techniques and required data sets to address these weaknesses. To do so, use established bias models and create necessary data sets. 
\item Research how annotators can be trained to understand media bias better and return better results in potential surveys and tasks. 
\item Research the perception of media bias from a psychological perspective to build a high-quality data set and evaluation. 
\item Implement a prototype of a media bias identification system that employs the developed methods to demonstrate the applicability of the approach in real-world news article collections. The target group of the prototype are non-expert people.
\item Evaluate the effectiveness of the bias identification methods with a test corpus.
\end{enumerate}

\section{Related Work}\label{sec:relatedwork}
Media bias can be defined as slanted news coverage or internal news article bias \cite{recasens2013a}. While there are multiple forms of bias, e.g., bias by personal perception or by the omission of information \cite{PUGLISI2015647}, our focus is on bias caused by word choice, in which different words refer to the same concept. I have published a detailed description of media bias in \cite{Spinde2021}, which I will refer to for more information. In the following, I summarize the existing literature on bias data sets and media bias classification. 

\subsection{Media Bias Data Sets}\label{sec:data sets}
\cite{lim2018b} present 1,235 sentences labeled for word and sentence level bias by crowdsourcing workers. All the sentences in their data set focus on one event. Another data set focusing on just one event is presented by \cite{10.1145/3340531.3412876}. It consists of 2,057 sentences from 90 news articles, annotated with bias labels on article and sentence levels, and contains labels such as overall bias, hidden assumption, and framing. The annotators agree with Krippendorff's $\alpha$ = -0.05. \cite{lim2020annotating} also provide a second data set with 966 sentences labeled on the sentence level. However, their reported interrater-agreement (IRR) of Fleiss' Kappa on different topics averages at zero. 

\cite{baumer2015a} classify framing in political news. Using crowdsourcing, they label 74 news articles from eight US news outlets, collected from politics-specific RSS feeds on two separate days. \cite{chen_analyzing_2020} create a data set of 6,964 articles containing political bias, unfairness, and non-objectivity labels at the article level. Altogether, they present 11 different topics such as "presidential election", "politics", and "white house".

\cite{fan2019a} present 300 news articles containing annotations for lexical and informational bias made by two experts. They define lexical bias as bias stemming from specific word choice, and informational bias as sentences conveying information tangential or speculative to sway readers' opinions towards entities \cite{fan2019a}. Their data set, BASIL, allows for analysis at the token level and relative to the target, but only 448 sentences are available for lexical bias.

Under the name MBIC, I \cite{spinde2021mbic} extract 1,700 sentences from 1,000 news articles. Crowdsource workers then label bias and opinion on a word and sentence level using a survey platform that also surveyed the annotators' backgrounds. MBIC covers 14 different topics and yields a Fleiss' Kappa score of 0.21. 

Even though the referenced data sets contribute valuable resources to the media bias investigation, they still have significant drawbacks, such as (1) a small number of topics \cite{lim2018b, lim2020annotating}, (2) no annotations on the word level \cite{lim2018b}, (3) low inter-annotator agreement \cite{spinde2021mbic, lim2020annotating, baumer2015a, lim2018b}, and (4) no background check for its participants (except \cite{spinde2021mbic}). Also, some related papers focus on framing rather than on bias \cite{baumer2015a, fan2019a}, and results are only partially transferable. In this work, I address these weaknesses by gathering sentence-level annotations about bias by word choice over a balanced and broad range of topics. The annotations are made by trained expert annotators with a higher capability of identifying bias than crowdsource workers. 

\subsection{Media Bias Classification Systems}
Several studies tackle the automated detection of media bias \cite{hube2018detecting, spinde2020a, chen_analyzing_2020}. Most of them use manually created features to detect bias \cite{hube2018detecting}, and are based on traditional machine learning models \cite{Spinde2021}.

\cite{recasens2013a} identify sentence level bias in Wikipedia using supervised classification. They use a bias lexicon and a set of various linguistic features (e.g., assertive verbs, sentiment) with a logistic regression classifier, identifying bias-inducing words in a sentence. They also report that crowdsource workers struggle to identify bias words their classifier is able to detect. 

In my work in \cite{Spinde2021}, I develop a feature-based tool to detect bias-inducing words. I identify and evaluate a wide range of linguistic, lexical, and syntactic features serving as potential bias indicators. The final classifier returns an $F_{1}$-score of 0.43 and 0.79 AUC. I also point out the explanatory power of various feature-based approaches and the performance of their own model on the MBIC data set. Still, I conclude that Deep Learning models are promising alternatives for future work. 

\cite{hube2018detecting} propose a semi-automated approach to extract domain-related bias based on word embeddings properties. 
The authors combine bias words and linguistic features  (e.g., report verbs, assertive verbs) in a random forest classifier to detect sentence-level bias in Wikipedia.
They achieve an $F_{1}$-score of 0.69 on a newly created ground truth based on Conservapedia.\footnote{\url{https://conservapedia.com/Main_Page}, accessed on 2021-04-10.} In their following work, \cite{hube2019neural} propose a neural statement-level bias detection approach based on Wikipedia data. Using recurrent neural networks (RNNs) and different attention mechanisms, the authors achieve an $F_{1}$-score of 0.77, indicating a possible advantage of neural classifiers in the domain.
\cite{chen_analyzing_2020} train a RNN to classify article-level bias. They also conduct a reverse feature analysis and find that, at the word level, political bias correlates with categories such as negative emotion, anger, and affect.

To summarize, most approaches use manually created features, leading to lower performance and poor representation. The few existing contributions on neural models are based on naive data sets  (Section \ref{sec:data sets}). Therefore, I develop a neural classifier trained on improved data sets. One of them is already finished, BABE. My current system incorporates state-of-the-art models and improves their pre-training step through distant supervision \cite{tang2014,deriu2017}, allowing the model to learn bias-specific embeddings, thus improving its representation.

\subsection{Psychological Background}\label{sec:psychology}
In various research areas, text perception and particularly bias detection have been investigated. For example, the influence of biased reporting within traditional, citizen, and social media has been researched \cite{ardevol-abreu_effects_2017}. Other projects focus on the perception of media outlets as being hostile [20], the influence of user-related variables on the perceptions of bias \cite{lee_liberal_2005, gunther_broad_2006, gunther_who_2017, ho_role_2011, rojas_corrective_2010}, or the perception of bias in particular topics \cite{glynn_how_2014}. Other main interests in the existing research are topic-dependent text perception \cite{dalessio_experimental_2003}, user comments \cite{houston_influence_2011, yun_hostile_2018}, and visual features \cite{peng_same_2018}. 

Apart from political or communication studies and psychology, an increasing number of computer science publications focus on the automated detection of media bias or the related concepts of framing, and sentiment analysis \cite{Spinde2021, spinde2021mbic, spinde2020a, spinde_tassy_2021, recasens2013a, lim2018a}. 

Independent of the scientific field (e.g., Psychology, Computer Science), all the research mentioned above questions students, experts, or crowdsource workers about their perception of bias on a word, sentence, article, or image level. However, almost none report a detailed process description on how they create the respective evaluation surveys or choose the questions handed to the participants. Also, especially in the computer science studies, except one of my studies \cite{Spinde2021}, none ask for the personal background of participants. Still, as shown in some of the work from psychology and communication science \cite{spinde2021B}, the personal background seems crucial to understand how to interpret and use the collected feedback annotations. The data sets used in the various computer scientific approaches and projects do not reflect media bias's complexity. Instead, they primarily focus on technical elements rather than the problem itself. I believe that data quality and comparability play a crucial role in training any classifier. Therefore, I already published a common and reliably evaluated question set, which is another central element of my overall dissertation. Again, I will refer more to that in Section \ref{sec:methodology}.

\subsection{Research Gap}
\begin{figure*}
\centering
\twocolumn[\resizebox{\textwidth}{!}{\includegraphics{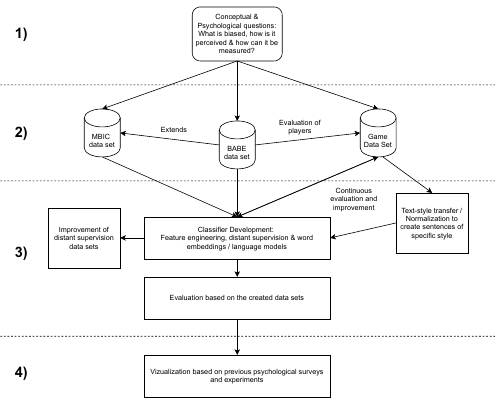}}
  \caption{Entire workflow of my dissertation project. Phase 1 refers to the interdisciplinary conceptual work, phase 2 to the data set creation, phase 3 to the classifier development and evaluation and phase 4 to the visualization of the results.}
  \label{workflow} \vspace{0.35cm}]
\end{figure*}

To my knowledge, all existing work is not including the psychological background of the concept in computer scientific work or vice-versa, based on data sets exhibiting weaknesses such as small size or no annotator background information and, therefore, return a measure of bias that is lacking transparency and reliability. Also, recent advancements in deep learning are not exploited within the media bias domain. To address the issues mentioned above, in \cref{workflow}, I propose a four-component architecture to solve my main research question. 
First (1)\footnote{Numbers refer to \cref{workflow}.}, I use the knowledge and models from sciences that have long studied media bias to conceptualize media bias overall and define possible measurements. Second (2), I build a more reliable (Annotator-background-aware) and bigger data set (or multiple data sets) that reflect media bias. Third (3), I develop a media bias classifier to identify bias automatically. Within this phase, I make use of the recent advancements of deep learning, including neural language models such as BERT \cite{devlin2018}. I believe Transformer-based\footnote{A Transformer is a deep learning model that adopts the mechanism of attention, weighing the influence of different parts of the input data. I have shown the advantages of Transformer models for media bias detection \cite{SpindeBABE}.} models hold potential to the outcome of my project, which partially has already been shown in my existing work with BABE. Fourth (4), I plan to display my results to potential users using visualizations I evaluated within multiple surveys.

\section{Data Set Creation}\label{sec:data}

Since media bias by word choice rarely depends on context outside the sentences \cite{fan2019a}, we focused on gathering sentences only. Mainly my work compasses three different data set creation approaches: 

\subsection{MBIC}
MBIC (\textbf{M}edia \textbf{B}ias Annotation Data Set \textbf{I}ncluding Annotator \textbf{C}haracteristics) is the first available dataset about media bias reporting detailed information on annotator characteristics and their individual background. It was the first data set I created. Before MBIC, existing data sets did not control for the individual background of annotators, which may affect their assessment and represent critical information for contextualizing their annotations. In the paper, I present a matrix-based methodology to crowdsource such data using a self-developed annotation platform called TASSY\cite{spinde_tassy_2021}. MBIC  contains 1,700 statements representing various media bias instances. Ten crowdsource annotators reviewed the statements each and contain labels for media bias identification both on the word and sentence level. MBIC achieves an inter-annotator agreement of $\alpha$ = 0.21. I also used it to train a feature-based media bias classifier \cite{Spinde2021}. 

\subsection{BABE}
 After my work on MBIC, I concluded that crowdsource workers struggle with understanding the concept of media bias, even when given detailed instructions. Therefore, I continued to develop BABE (\textbf{B}ias \textbf{A}nnotations \textbf{B}y \textbf{E}xperts) \cite{SpindeBABE}. BABE is a robust and diverse data set created by trained experts, for media bias research. In the paper, I  analyze why expert labeling is essential within this domain\footnote{The exact collection strategy and other details are laid out in \cite{SpindeBABE}.}. The data set offers better annotation quality and higher inter-annotator agreement than existing work. It consists of 3,700 sentences balanced among topics and outlets, containing media bias labels on the word and sentence level. Also, it allowed me to compare expert annotations with the crowdsourced labels provided by \cite{spinde2021mbic} to analyze quality differences between the two groups further. My results show how expert annotators render more qualitative bias labels than crowdsource workers in MBIC. Employing annotators with domain expertise allows me to achieve an inter-annotator agreement of $\alpha$ = 0.40, which is higher than existing data sets \cite{spinde2021mbic}. I believe domain knowledge and training alleviate the difficulty of identifying bias and are imperative to create a strong benchmark due to the complexity of the task. In future work, apart from improving the current data set and classifier, I will also explore why a text passage might be biased, not just its overall classification.
  
\subsection{Gamification}
While BABE offers a clear performance increase compared to MBIC, and I also showed that a high-quality media bias data set needs trained annotators, it contains only 3700 sentences, and already cost over 10.000\euro. 
Given the high costs for human annotation, I will propose a different paradigm to mitigate this problem. Gamification of the process will be used to obtain/produce the desired labels. The game will be hosted on an open platform like Zooniverse\footnote{\url{https://www.zooniverse.org/}, accessed on 2021-04-10.} and it will be promoted in universities/schools. I am designing it to be motivating, serious, and teach people to understand the complex construct of media bias. The game will be based on several layers; I will only summarize them here briefly: 
\begin{enumerate}
    \item In the first round, players will see a video and tutorial for sentence-level bias annotation. They will also be instructed about the game mechanics and ranking system. 
     \item In the second round, players will annotate sentences from BABE and get direct feedback whether or not they agree with the expert annotations made there. 
      \item From the third round on, players will annotate new sentences. They will be evaluated and given feedback, dependent on how other players rate the sentences. 
       \item From the fourth round on, players will also be able to write sentences that others can evaluate to experiment with different writing styles. 
\end{enumerate}
A smaller study is currently in execution to design the game properly, researching ideal ways to teach readers about bias and critical reading in general. The study contains various bias visualizations, an inoculation message, and a video about bias. 


\section{Classifier Methodology}\label{sec:methodology}
After some preliminary work \cite{Spinde2021, spinde2020a, spinde2020b, Spinde2020INRA}, I propose the use of neural classifiers with automated feature learning capabilities to solve the given media bias classification task. A distant supervision framework, similar to \cite{howard2018}, allows me to pre-train the feature extraction algorithms leading to improved language representations, thus, including information about bias in any sample. As obtaining large amounts of pre-training labeled data using humans is prohibitively expensive, I resort to noisy yet abundantly available labels providing supervisory signals within the BABE paper. I will provide a high-quality distant supervision data set for the overall project. However, the approach used with BABE already shows that distant supervision is a promising approach to tackle media bias identification. 

\subsection{Learning Task}
Given a corpus $X$ and a randomly sampled sequence of tokens $x_i \in X$ with $i \in \{1,...,N\}$, the learning task in such an approach consists of assigning the correct label $y_i$ to $x_i$ where $y_i \in \{0,1\}$ represents the \textit{neutral} and \textit{biased} classes, respectively. The supervised task can be optimized by minimizing the binary cross-entropy loss
\begin{equation}
\label{eq:loss}
\mathcal{L} := - \frac{1}{N} \sum_{i=1}^{N} \sum_{k=\{0,1\}} f_k(x_i) \cdot log(\hat{f_k}(x_i)).
\end{equation}
where $f_k(\cdot)$ is a binary indicator triggering 0 in the case of neutral labels and 1 in the case of a biased sequence. $\hat{f}_k(\cdot)$ is a scalar representing the language model score for the given sequence. 

\subsection{Neural Models}
Overall, I fit $\hat{f}_k(\cdot)$ using a range of state-of-the-art language models. Central to the architectural design of these models is \cite{vaswani2017} 's encoder stack of the Transformer\cite{vaswani2017} relying solely on the attention mechanism. I use the BERT model \cite{devlin2018} and its variants DistilBERT \cite{sanh2019} and RoBERTa \cite{liu2019} that learned bidirectional language representations from the unlabeled text. DistilBERT is a compressed model of the original BERT, and RoBERTa uses a slightly different loss function with more training data than its predecessor. I also evaluate models built on the transformer architecture but differ in the training objective. While DistilBERT and RoBERTa use masked language modeling as a pre-training task, ELECTRA \cite{clark2020} uses a discriminative approach to learn language representations. I also include XLNet \cite{yang2019} in our comparison as an example of an autoregressive model. I systematically evaluate the performance of the model on the media bias sentence classification task and investigate the impact of an additional pre-training task introduced in the next section on the classification capabilities of the models BERT and RoBERTa.

\subsection{Distant Supervision}
Fine-tuning general language models on the target task has proven beneficial for many tasks in NLP \cite{SpindeBABE}. The language model pre-training followed by fine-tuning allows models to incorporate the idiosyncrasies of the target corpus. For text classification, the authors of ULMFiT \cite{howard2018} demonstrated the superiority of task-specific word embeddings. Before fine-tuning, I introduce an additional pre-training task to improve feature learning capabilities considering media bias content. 
The typical unsupervised setting used in the general pre-training stage does not include information on language bias in the learning of the embedded space. To remedy this, I incorporate bias information directly in the loss function (equation \ref{eq:loss}) via distant supervision. In this approach, distant or \textit{weak} labels are predicted from noisy sources, alleviating the need for data labeled by humans. Results by \cite{severyn2015} and \cite{deriu2017} demonstrated that pre-training on larger distant datasets followed by fine-tuning on supervised data yields improved performance for sentiment classification.

\subsection{Distant Supervision Data Set}
Until now, my pre-training corpus to apply distant supervision comprises news headlines of outlets with and without a partisan leaning to learn bias-specific word embeddings. The data source, various news outlets, are leveraged to provide distant supervision to our system. As a result, the large amounts of data necessary to learn continuous word representations are gathered by mechanical means alleviating the burden of collecting expensive annotations. The assumption is that the distribution of biased words is denser in some news sources than in others. Text sampled from news outlets with a partisan leaning\footnote{The leaning bases on the Media Bias Chart on \url{https://www.allsides.com/media-bias/media-bias-chart}, accessed on 2021-04-13.}  is treated as biased. Text sampled from news organizations with high journalistic standards is treated as neutral. Thus, the mapping of bias and neutral labels to sequences is automatized. The data collection resembles the collection of the ground-truth data described in Section \ref{sec:data}. The defined keywords reflect contentious issues of the US society, as I assume slanted reporting to be more likely among those topics than in the case of less controversial topics. The obtained corpus consisting of 83,143 neutral news headlines and 45,605 biased instances allows for the encoding of a sequence's bias information in the embedded space. The news headlines corpus serves to learn more effective language representations; it is unsuitable for evaluation purposes due to its noisy nature. I ensure  no overlap exists between the distant corpus and BABE to guarantee model integrity with respect to training and testing.

As a next project step, I am testing various distant data sets based on polarizing content from Twitter, Reddit, Wikipedia, and manually created selections of biased on non-biased content. I have already shown that some of these data sets improve bias detection tasks within a multi-task-learning approach \cite{SpindeMTL}. Also, I am developing a parallel data set of neural and biased sentences together with a text style transfer system to experiment with automatically creating a distant supervision data set. Ideally, the system should automatically adjust and improve its performance: When sentences of different styles can be found or generated while the game data set returns manually created evaluations, performance can be continuously adjusted on an automated scale. 


\section{System and Visualization}\label{sec:visualization}
A system will integrate the previously described analysis workflow and visualize the results to non-expert users. I devised visualizations similar to UIs of popular news aggregators, such as Google News, and bias-aware aggregators, such as AllSides. In contrast to these, the system will be able to identify in-text instances of bias \cite{spinde2020b}. Hence, the system will give a bias-aware overview of current topics and have a visualization for single articles, which will highlight identified instances of bias. For research and evaluation of the previously described system and its analysis methods, I currently use a data set incorporating the AllSides labels \cite{chen_analyzing_2020} and BABE, which have high diversity concerning the political slant of outlets. I already researched how bias visualizations can work for everyday users in multiple studies, which I will continue to expand \cite{spinde2020b}. These projects aim to contribute to a deeper understanding of effective media bias communication. To this end, I create a set of bias visualizations revealing bias in different ways and test their effectiveness in online experiments \cite{spinde2020b}.


\section{Evaluation}\label{sec:evaluation}
So far, all my work is evaluated in a traditional way by the binary biased/non-biased labels returned in studies and other related data sets. Still, I argue that standard metrics (e.g., accuracy and $F_1$) provide a limited perspective into a model's predictive power in case of a complex construct like media bias. Further research needs to tackle these pitfalls to propose systems with better generalization capabilities. A promising starting point might be a more refined evaluation scheme that decomposes the bias detection task into multiple sub-tasks, such as presented in CheckList \cite{ribeiro2020beyond}. This scheme allows us to understand how our system performs on different types of bias (e.g., bias by context, linguistics, and overall reporting). My final bias classifier will not be evaluated on a binary scale but on a multi-faceted data set containing other related concepts like hate speech, sentiment, polarity, linguistic structure and figures, and personal annotator background per annotation. The respective data set is currently in progress. 

Additionally, I believe that current research on explainable artificial intelligence might increase user trust in classifiers. Existing research presents ways to visualize Transformer-based models and make their results more accessible and interpretable \cite{vig2019}. Lastly, combining neural methods with advances in linguistic bias theory \cite{Spinde2021} to explain a classifier's decision to users will also be part of my ongoing work, which I already inspected in one feature-based approach \cite{Spinde2021}.

I mostly focus on sentence-level bias, which is often used in the media bias domain, but also include word-level annotations in every data set to encourage solutions focusing on more granular characteristics. I believe word-level bias conveys a strong explanatory power and is, therefore, a promising research direction. I will also address the word level in future work. 

\section{Conclusion and Implications}\label{sec:conclusion}
In summary, both everyday news consumers and researchers could benefit from the automated identification of bias in news articles. Devising suitable methods and data sets to find, automatically identify, and visualize bias are at the heart of this research project. The currently finished classifier based on BABE outperforms existing classifiers in the area. In addition, BABE achieves higher quality and annotator agreement than existing data sets \cite{lim2018b, lim2020annotating, baumer2015a,fan2019a}. Also, the media bias game will be rolled out into schools and Zooniverse to teach a broad audience about bias and return a high amount of annotations simultaneously. My vision is that at a later point in time, the developed methods will make their way into an actual app or tool, helping news readers to explore and understand media bias through their daily news consumption. I also believe my system will provide promising grounds for large-scale bias analyzes as they are already executed in the social sciences or economics.

\section*{Acknowledgment}
I want to thank the Hanns-Seidel-Foundation, Germany, which supports this work.



\bibliographystyle{IEEEtran}
\bibliography{_custom.bib}

\begin{thebibliography}{10}
\providecommand{\url}[1]{#1}
\csname url@samestyle\endcsname
\providecommand{\newblock}{\relax}
\providecommand{\bibinfo}[2]{#2}
\providecommand{\BIBentrySTDinterwordspacing}{\spaceskip=0pt\relax}
\providecommand{\BIBentryALTinterwordstretchfactor}{4}
\providecommand{\BIBentryALTinterwordspacing}{\spaceskip=\fontdimen2\font plus
\BIBentryALTinterwordstretchfactor\fontdimen3\font minus
  \fontdimen4\font\relax}
\providecommand{\BIBforeignlanguage}[2]{{%
\expandafter\ifx\csname l@#1\endcsname\relax
\typeout{** WARNING: IEEEtran.bst: No hyphenation pattern has been}%
\typeout{** loaded for the language `#1'. Using the pattern for}%
\typeout{** the default language instead.}%
\else
\language=\csname l@#1\endcsname
\fi
#2}}
\providecommand{\BIBdecl}{\relax}
\BIBdecl

\bibitem{dallmann2015a}
\BIBentryALTinterwordspacing
A.~Dallmann, F.~Lemmerich, D.~Zoller, and A.~Hotho, ``Media bias in german
  online newspapers,'' in \emph{Proceedings of the 26th ACM Conference on
  Hypertext \& Social Media}, ser. HT '15.\hskip 1em plus 0.5em minus
  0.4em\relax New York, NY, USA: Association for Computing Machinery, 2015, p.
  133–137. [Online]. Available: \url{https://doi.org/10.1145/2700171.2791057}
\BIBentrySTDinterwordspacing

\bibitem{Spinde2021}
\BIBentryALTinterwordspacing
T.~Spinde, L.~Rudnitckaia, J.~Mitrovi{\'{c}}, F.~Hamborg, M.~Granitzer,
  B.~Gipp, and K.~Donnay, ``{Automated identification of bias inducing words in
  news articles using linguistic and context-oriented features},''
  \emph{Information Processing {\&} Management}, vol.~58, no.~3, p. 102505,
  2021. [Online]. Available: \url{https://doi.org/10.1016/j.ipm.2021.102505}
\BIBentrySTDinterwordspacing

\bibitem{spinde2021mbic}
T.~Spinde, L.~Rudnitckaia, K.~Sinha, F.~Hamborg, B.~Gipp, and K.~Donnay,
  ``{MBIC} -- {A} media bias annotation dataset including annotator
  characteristics,'' in \emph{Proceedings of the iConference 2021}.\hskip 1em
  plus 0.5em minus 0.4em\relax iSchools, 2021.

\bibitem{spinde_how_2021}
T.~Spinde, C.~Kreuter, W.~Gaissmaier, F.~Hamborg, B.~Gipp, and H.~Giese, ``Do
  you think it's biased? how to ask for the perception of media bias,'' in
  \emph{Proceedings of the {ACM}/{IEEE} {Joint} {Conference} on {Digital}
  {Libraries} ({JCDL})}, Sep. 2021.

\bibitem{spinde2020b}
\BIBentryALTinterwordspacing
T.~Spinde, F.~Hamborg, K.~Donnay, A.~Becerra, and B.~Gipp, ``Enabling news
  consumers to view and understand biased news coverage: A study on the
  perception and visualization of media bias,'' in \emph{Proceedings of the
  ACM/IEEE Joint Conference on Digital Libraries in 2020}, ser. JCDL '20.\hskip
  1em plus 0.5em minus 0.4em\relax Virtual Event, China: Association for
  Computing Machinery, 2020, p. 389–392. [Online]. Available:
  \url{https://doi.org/10.1145/3383583.3398619}
\BIBentrySTDinterwordspacing

\bibitem{SpindeOmission}
\BIBentryALTinterwordspacing
J.~Ehrhardt, T.~Spinde, A.~Vardasbi, and F.~Hamborg, ``Omission of information:
  Identifying political slant via an analysis of co-occurring entities,'' in
  \emph{Information between Data and Knowledge}, ser. Schriften zur
  Informationswissenschaft.\hskip 1em plus 0.5em minus 0.4em\relax
  Gl{\"u}ckstadt: Werner H{\"u}lsbusch, 2021, vol.~74, pp. 80--93, session 2:
  Information Behavior and Information Literacy 2. [Online]. Available:
  \url{https://epub.uni-regensburg.de/44939/}
\BIBentrySTDinterwordspacing

\bibitem{park2009newscube}
S.~Park, S.~Kang, S.~Chung, and J.~Song, ``\BIBforeignlanguage{en}{{NewsCube}:
  delivering multiple aspects of news to mitigate media bias},'' in
  \emph{\BIBforeignlanguage{en}{Proceedings of the 27th international
  conference on {Human} factors in computing systems - {CHI} 09}}.\hskip 1em
  plus 0.5em minus 0.4em\relax Boston, MA, USA: ACM Press, 2009, p. 443.

\bibitem{baumer2015a}
\BIBentryALTinterwordspacing
E.~Baumer, E.~Elovic, Y.~Qin, F.~Polletta, and G.~Gay, ``Testing and comparing
  computational approaches for identifying the language of framing in political
  news,'' in \emph{Proceedings of the 2015 Conference of the North {A}merican
  Chapter of the Association for Computational Linguistics: Human Language
  Technologies}.\hskip 1em plus 0.5em minus 0.4em\relax Denver, Colorado:
  Association for Computational Linguistics, May{--}Jun. 2015, pp. 1472--1482.
  [Online]. Available: \url{https://www.aclweb.org/anthology/N15-1171}
\BIBentrySTDinterwordspacing

\bibitem{Spinde2021Embeddings}
T.~Spinde, L.~Rudnitckaia, F.~Hamborg, and B.~Gipp, ``Identification of biased
  terms in news articles by comparison of outlet-specific word embeddings,'' in
  \emph{Proceedings of the iConference 2021}, March 2021.

\bibitem{recasens2013a}
\BIBentryALTinterwordspacing
M.~Recasens, C.~Danescu-Niculescu-Mizil, and D.~Jurafsky, ``Linguistic models
  for analyzing and detecting biased language,'' in \emph{Proceedings of the
  51st Annual Meeting of the Association for Computational Linguistics (Volume
  1: Long Papers)}, 2013, pp. 1650--1659. [Online]. Available:
  \url{https://www.aclweb.org/anthology/P13-1162.pdf}
\BIBentrySTDinterwordspacing

\bibitem{lim2020annotating}
\BIBentryALTinterwordspacing
S.~Lim, A.~Jatowt, M.~F{\"a}rber, and M.~Yoshikawa,
  ``\BIBforeignlanguage{English}{Annotating and analyzing biased sentences in
  news articles using crowdsourcing},'' in
  \emph{\BIBforeignlanguage{English}{Proceedings of the 12th Language Resources
  and Evaluation Conference}}.\hskip 1em plus 0.5em minus 0.4em\relax
  Marseille, France: European Language Resources Association, May 2020, pp.
  1478--1484. [Online]. Available:
  \url{https://www.aclweb.org/anthology/2020.lrec-1.184}
\BIBentrySTDinterwordspacing

\bibitem{hube2018detecting}
\BIBentryALTinterwordspacing
C.~Hube and B.~Fetahu, ``Detecting biased statements in wikipedia,'' in
  \emph{Companion Proceedings of the The Web Conference 2018}, ser. WWW
  '18.\hskip 1em plus 0.5em minus 0.4em\relax Republic and Canton of Geneva,
  CHE: International World Wide Web Conferences Steering Committee, 2018, p.
  1779–1786. [Online]. Available:
  \url{https://doi.org/10.1145/3184558.3191640}
\BIBentrySTDinterwordspacing

\bibitem{hube2019neural}
\BIBentryALTinterwordspacing
------, ``Neural based statement classification for biased language,'' in
  \emph{Proceedings of the Twelfth ACM International Conference on Web Search
  and Data Mining}, ser. WSDM '19.\hskip 1em plus 0.5em minus 0.4em\relax New
  York, NY, USA: Association for Computing Machinery, 2019, p. 195–203.
  [Online]. Available: \url{https://doi.org/10.1145/3289600.3291018}
\BIBentrySTDinterwordspacing

\bibitem{chen_analyzing_2020}
\BIBentryALTinterwordspacing
W.-F. Chen, K.~Al~Khatib, H.~Wachsmuth, and B.~Stein,
  ``\BIBforeignlanguage{en}{Analyzing {Political} {Bias} and {Unfairness} in
  {News} {Articles} at {Different} {Levels} of {Granularity}},'' in
  \emph{\BIBforeignlanguage{en}{Proceedings of the {Fourth} {Workshop} on
  {Natural} {Language} {Processing} and {Computational} {Social}
  {Science}}}.\hskip 1em plus 0.5em minus 0.4em\relax Online: Association for
  Computational Linguistics, 2020, pp. 149--154. [Online]. Available:
  \url{https://www.aclweb.org/anthology/2020.nlpcss-1.16}
\BIBentrySTDinterwordspacing

\bibitem{gunther_assimilation_2009}
\BIBentryALTinterwordspacing
A.~C. Gunther, N.~Miller, and J.~L. Liebhart,
  ``\BIBforeignlanguage{en}{Assimilation and {Contrast} in a {Test} of the
  {Hostile} {Media} {Effect}:},'' \emph{\BIBforeignlanguage{en}{Communication
  Research}}, Oct. 2009. [Online]. Available:
  \url{https://journals.sagepub.com/doi/10.1177/0093650209346804}
\BIBentrySTDinterwordspacing

\bibitem{kim_partisans_2015}
\BIBentryALTinterwordspacing
M.~Kim, ``Partisans and {Controversial} {News} {Online}: {Comparing}
  {Perceptions} of {Bias} and {Credibility} in {News} {Content} {From} {Blogs}
  and {Mainstream} {Media},'' \emph{Mass Communication and Society}, vol.~18,
  no.~1, pp. 17--36, Jan. 2015. [Online]. Available:
  \url{https://doi.org/10.1080/15205436.2013.877486}
\BIBentrySTDinterwordspacing

\bibitem{kim_role_2016}
\BIBentryALTinterwordspacing
{M. Kim}, ``The {Role} of {Partisan} {Sources} and {Audiences}' {Involvement}
  in {Bias} {Perceptions} of {Controversial} {News},'' \emph{Media Psychology},
  vol.~19, no.~2, pp. 203--223, Apr. 2016. [Online]. Available:
  \url{https://doi.org/10.1080/15213269.2014.1002941}
\BIBentrySTDinterwordspacing

\bibitem{lee_liberal_2005}
\BIBentryALTinterwordspacing
T.-T. Lee, ``The {Liberal} {Media} {Myth} {Revisited}: {An} {Examination} of
  {Factors} {Influencing} {Perceptions} of {Media} {Bias},'' \emph{Journal of
  Broadcasting \& Electronic Media}, vol.~49, no.~1, pp. 43--64, 2005,
  publisher: Routledge. [Online]. Available:
  \url{https://doi.org/10.1207/s15506878jobem4901\_4}
\BIBentrySTDinterwordspacing

\bibitem{gunther_broad_2006}
A.~C. Gunther and J.~L. Liebhart, ``\BIBforeignlanguage{en}{Broad {Reach} or
  {Biased} {Source}? {Decomposing} the {Hostile} {Media} {Effect}},''
  \emph{\BIBforeignlanguage{en}{Journal of Communication}}, vol.~56, no.~3, pp.
  449--466, Sep. 2006.

\bibitem{glynn_how_2014}
\BIBentryALTinterwordspacing
C.~J. Glynn and M.~E. Huge, ``\BIBforeignlanguage{en}{How {Pervasive} {Are}
  {Perceptions} of {Bias}? {Exploring} {Judgments} of {Media} {Bias} in
  {Financial} {News}},'' \emph{\BIBforeignlanguage{en}{International Journal of
  Public Opinion Research}}, vol.~26, no.~4, pp. 543--553, Dec. 2014. [Online].
  Available: \url{https://academic.oup.com/ijpor/article/26/4/543/734722}
\BIBentrySTDinterwordspacing

\bibitem{houston_influence_2011}
\BIBentryALTinterwordspacing
J.~B. Houston, G.~J. Hansen, and G.~S. Nisbett,
  ``\BIBforeignlanguage{en}{Influence of {User} {Comments} on {Perceptions} of
  {Media} {Bias} and {Third}-{Person} {Effect} in {Online} {News}},''
  \emph{\BIBforeignlanguage{en}{Electronic News}}, vol.~5, no.~2, pp. 79--92,
  Jun. 2011. [Online]. Available:
  \url{https://doi.org/10.1177/1931243111407618}
\BIBentrySTDinterwordspacing

\bibitem{spinde2020a}
\BIBentryALTinterwordspacing
T.~Spinde, F.~Hamborg, and B.~Gipp, ``An integrated approach to detect media
  bias in german news articles,'' in \emph{Proceedings of the ACM/IEEE Joint
  Conference on Digital Libraries in 2020}, ser. JCDL '20.\hskip 1em plus 0.5em
  minus 0.4em\relax Virtual Event, China: Association for Computing Machinery,
  2020, p. 505–506. [Online]. Available:
  \url{https://doi.org/10.1145/3383583.3398585}
\BIBentrySTDinterwordspacing

\bibitem{Spinde2020INRA}
\BIBentryALTinterwordspacing
------, ``Media bias in german news articles: A combined approach,'' \emph{ECML
  PKDD 2020 Workshops: Workshops of the European Conference on Machine Learning
  and Knowledge Discovery in Databases (ECML PKDD 2020): INRA 2020, Ghent,
  Belgium, September 14--18, 2020, Proceedings}, vol. 1323, pp. 581--590, 2020.
  [Online]. Available:
  \url{https://www.ncbi.nlm.nih.gov/pmc/articles/PMC7850083/}
\BIBentrySTDinterwordspacing

\bibitem{spinde_towards_2021}
T.~Spinde, D.~Krieger, M.~Plank, and B.~Gipp, ``Towards {A} {Reliable}
  {Ground}-{Truth} {For} {Biased} {Language} {Detection},'' in
  \emph{Proceedings of the {ACM}/{IEEE} {Joint} {Conference} on {Digital}
  {Libraries} ({JCDL})}, Sep. 2021.

\bibitem{spinde2021B}
T.~Spinde, C.~Kreuter, W.~Gaissmaier, F.~Hamborg, B.~Gipp, and H.~Giese,
  ``\BIBforeignlanguage{en}{How can the perception of media bias in news
  articles be objectively measured? {B}est practices and recommendations using
  user studies.}'' in \emph{\BIBforeignlanguage{en}{Proceedings of the ACM/IEEE
  Joint Conference on Digital Libraries in 2021 [in review]}}, ser. JCDL
  '21.\hskip 1em plus 0.5em minus 0.4em\relax Virtual Event, US: Association
  for Computing Machinery, 2021.

\bibitem{PUGLISI2015647}
\BIBentryALTinterwordspacing
R.~Puglisi and J.~M. Snyder, ``Empirical studies of media bias,'' in
  \emph{Handbook of Media Economics}, ser. Handbook of Media Economics, S.~P.
  Anderson, J.~Waldfogel, and D.~Strömberg, Eds.\hskip 1em plus 0.5em minus
  0.4em\relax North-Holland, 2015, vol.~1, pp. 647--667. [Online]. Available:
  \url{https://www.sciencedirect.com/science/article/pii/B9780444636850000152}
\BIBentrySTDinterwordspacing

\bibitem{lim2018b}
\BIBentryALTinterwordspacing
S.~Lim, A.~Jatowt, and M.~Yoshikawa, ``Understanding characteristics of biased
  sentences in news articles,'' in \emph{CIKM Workshops}, 2018. [Online].
  Available: \url{http://ceur-ws.org/Vol-2482/paper13.pdf}
\BIBentrySTDinterwordspacing

\bibitem{10.1145/3340531.3412876}
\BIBentryALTinterwordspacing
M.~F\"{a}rber, V.~Burkard, A.~Jatowt, and S.~Lim, ``A multidimensional dataset
  based on crowdsourcing for analyzing and detecting news bias,'' in
  \emph{Proceedings of the 29th ACM International Conference on Information \&
  Knowledge Management}, ser. CIKM '20.\hskip 1em plus 0.5em minus 0.4em\relax
  New York, NY, USA: Association for Computing Machinery, 2020, p. 3007–3014.
  [Online]. Available: \url{https://doi.org/10.1145/3340531.3412876}
\BIBentrySTDinterwordspacing

\bibitem{fan2019a}
\BIBentryALTinterwordspacing
L.~Fan, M.~White, E.~Sharma, R.~Su, P.~K. Choubey, R.~Huang, and L.~Wang, ``In
  plain sight: Media bias through the lens of factual reporting,'' in
  \emph{Proceedings of the 2019 Conference on Empirical Methods in Natural
  Language Processing and the 9th International Joint Conference on Natural
  Language Processing (EMNLP-IJCNLP)}.\hskip 1em plus 0.5em minus 0.4em\relax
  Hong Kong, China: Association for Computational Linguistics, Nov. 2019, pp.
  6343--6349. [Online]. Available:
  \url{https://www.aclweb.org/anthology/D19-1664}
\BIBentrySTDinterwordspacing

\bibitem{tang2014}
\BIBentryALTinterwordspacing
D.~Tang, F.~Wei, N.~Yang, M.~Zhou, T.~Liu, and B.~Qin, ``Learning
  sentiment-specific word embedding for {T}witter sentiment classification,''
  in \emph{Proceedings of the 52nd Annual Meeting of the Association for
  Computational Linguistics (Volume 1: Long Papers)}.\hskip 1em plus 0.5em
  minus 0.4em\relax Baltimore, Maryland: Association for Computational
  Linguistics, Jun. 2014, pp. 1555--1565. [Online]. Available:
  \url{https://www.aclweb.org/anthology/P14-1146}
\BIBentrySTDinterwordspacing

\bibitem{deriu2017}
J.~Deriu, A.~Lucchi, V.~D. Luca, A.~Severyn, S.~Müller, M.~Cieliebak,
  T.~Hofmann, and M.~Jaggi, ``Leveraging large amounts of weakly supervised
  data for multi-language sentiment classification,'' 2017.

\bibitem{ardevol-abreu_effects_2017}
\BIBentryALTinterwordspacing
A.~Ardèvol-Abreu and H.~Gil~de Zúñiga, ``\BIBforeignlanguage{en}{Effects of
  {Editorial} {Media} {Bias} {Perception} and {Media} {Trust} on the {Use} of
  {Traditional}, {Citizen}, and {Social} {Media} {News}},''
  \emph{\BIBforeignlanguage{en}{Journalism \& Mass Communication Quarterly}},
  vol.~94, no.~3, pp. 703--724, Sep. 2017. [Online]. Available:
  \url{https://doi.org/10.1177/1077699016654684}
\BIBentrySTDinterwordspacing

\bibitem{gunther_who_2017}
\BIBentryALTinterwordspacing
A.~C. Gunther, B.~McLaughlin, M.~R. Gotlieb, and D.~Wise,
  ``\BIBforeignlanguage{en}{Who {Says} {What} to {Whom}: {Content} {Versus}
  {Source} in the {Hostile} {Media} {Effect}},''
  \emph{\BIBforeignlanguage{en}{International Journal of Public Opinion
  Research}}, vol.~29, no.~3, pp. 363--383, Sep. 2017. [Online]. Available:
  \url{https://academic.oup.com/ijpor/article/29/3/363/2669466}
\BIBentrySTDinterwordspacing

\bibitem{ho_role_2011}
\BIBentryALTinterwordspacing
S.~S. Ho, A.~R. Binder, A.~B. Becker, P.~Moy, D.~A. Scheufele, D.~Brossard, and
  A.~C. Gunther, ``The {Role} of {Perceptions} of {Media} {Bias} in {General}
  and {Issue}-{Specific} {Political} {Participation},'' \emph{Mass
  Communication and Society}, vol.~14, no.~3, pp. 343--374, May 2011. [Online].
  Available: \url{https://doi.org/10.1080/15205436.2010.491933}
\BIBentrySTDinterwordspacing

\bibitem{rojas_corrective_2010}
\BIBentryALTinterwordspacing
H.~Rojas, ``\BIBforeignlanguage{en}{“{Corrective}” {Actions} in the
  {Public} {Sphere}: {How} {Perceptions} of {Media} and {Media} {Effects}
  {Shape} {Political} {Behaviors}},''
  \emph{\BIBforeignlanguage{en}{International Journal of Public Opinion
  Research}}, vol.~22, no.~3, pp. 343--363, Oct. 2010. [Online]. Available:
  \url{https://academic.oup.com/ijpor/article/22/3/343/691318}
\BIBentrySTDinterwordspacing

\bibitem{dalessio_experimental_2003}
\BIBentryALTinterwordspacing
D.~D'Alessio, ``\BIBforeignlanguage{en}{An {Experimental} {Examination} of
  {Readers}' {Perceptions} of {Media} {Bias}},''
  \emph{\BIBforeignlanguage{en}{Journalism \& Mass Communication Quarterly}},
  vol.~80, no.~2, pp. 282--294, Jun. 2003. [Online]. Available:
  \url{https://doi.org/10.1177/107769900308000204}
\BIBentrySTDinterwordspacing

\bibitem{yun_hostile_2018}
\BIBentryALTinterwordspacing
G.~W. Yun, S.-Y. Park, S.~Lee, and M.~A. Flynn, ``Hostile {Media} or {Hostile}
  {Source}? {Bias} {Perception} of {Shared} {News},'' \emph{Social Science
  Computer Review}, vol.~36, no.~1, pp. 21--35, Feb. 2018. [Online]. Available:
  \url{https://doi.org/10.1177/0894439316684481}
\BIBentrySTDinterwordspacing

\bibitem{peng_same_2018}
\BIBentryALTinterwordspacing
Y.~Peng, ``\BIBforeignlanguage{en}{Same {Candidates}, {Different} {Faces}:
  {Uncovering} {Media} {Bias} in {Visual} {Portrayals} of {Presidential}
  {Candidates} with {Computer} {Vision}},''
  \emph{\BIBforeignlanguage{en}{Journal of Communication}}, vol.~68, no.~5, pp.
  920--941, Oct. 2018. [Online]. Available:
  \url{https://academic.oup.com/joc/article/68/5/920/5113150}
\BIBentrySTDinterwordspacing

\bibitem{spinde_tassy_2021}
T.~Spinde, K.~Sinha, N.~Meuschke, and B.~Gipp, ``{TASSY} - {A} {Text}
  {Annotation} {Survey} {System},'' in \emph{Proceedings of the {ACM}/{IEEE}
  {Joint} {Conference} on {Digital} {Libraries} ({JCDL})}, Sep. 2021.

\bibitem{lim2018a}
S.~Lim, A.~Jatowt, and M.~Yoshikawa, ``\BIBforeignlanguage{en}{Deim forum 2018
  c1-3 towards bias inducing word detection by linguistic cue analysis in news
  articles},'' 2018.

\bibitem{devlin2018}
\BIBentryALTinterwordspacing
J.~Devlin, M.-W. Chang, K.~Lee, and K.~Toutanova, ``{BERT}: Pre-training of
  deep bidirectional transformers for language understanding,'' in
  \emph{Proceedings of the 2019 Conference of the North {A}merican Chapter of
  the Association for Computational Linguistics: Human Language Technologies,
  Volume 1 (Long and Short Papers)}.\hskip 1em plus 0.5em minus 0.4em\relax
  Minneapolis, Minnesota: Association for Computational Linguistics, Jun. 2019,
  pp. 4171--4186. [Online]. Available:
  \url{https://www.aclweb.org/anthology/N19-1423}
\BIBentrySTDinterwordspacing

\bibitem{SpindeBABE}
T.~Spinde, M.~Plank, J.-D. Krieger, T.~Ruas, B.~Gipp, and A.~Aizawa, ``{N}eural
  {M}edia {B}ias {D}etection {U}sing {D}istant {S}upervision {W}ith {BABE} -
  {B}ias {A}nnotations {B}y {E}xperts,'' in \emph{Findings of the Association
  for Computational Linguistics: EMNLP 2021}, Dominican Republic, Nov. 2021.

\bibitem{howard2018}
\BIBentryALTinterwordspacing
J.~Howard and S.~Ruder, ``Universal language model fine-tuning for text
  classification,'' in \emph{Proceedings of the 56th Annual Meeting of the
  Association for Computational Linguistics (Volume 1: Long Papers)}.\hskip 1em
  plus 0.5em minus 0.4em\relax Melbourne, Australia: Association for
  Computational Linguistics, Jul. 2018, pp. 328--339. [Online]. Available:
  \url{https://www.aclweb.org/anthology/P18-1031}
\BIBentrySTDinterwordspacing

\bibitem{vaswani2017}
\BIBentryALTinterwordspacing
A.~Vaswani, N.~Shazeer, N.~Parmar, J.~Uszkoreit, L.~Jones, A.~N. Gomez, L.~u.
  Kaiser, and I.~Polosukhin, ``Attention is all you need,'' in \emph{Advances
  in Neural Information Processing Systems}, I.~Guyon, U.~V. Luxburg,
  S.~Bengio, H.~Wallach, R.~Fergus, S.~Vishwanathan, and R.~Garnett, Eds.,
  vol.~30.\hskip 1em plus 0.5em minus 0.4em\relax Curran Associates, Inc.,
  2017. [Online]. Available:
  \url{\url{https://proceedings.neurips.cc/paper/2017/file/3f5ee243547dee91fbd053c1c4a845aa-Paper.pdf}}
\BIBentrySTDinterwordspacing

\bibitem{sanh2019}
\BIBentryALTinterwordspacing
V.~Sanh, L.~Debut, J.~Chaumond, and T.~Wolf, ``Distilbert, a distilled version
  of {BERT:} smaller, faster, cheaper and lighter,'' \emph{CoRR}, vol.
  abs/1910.01108, 2019. [Online]. Available:
  \url{http://arxiv.org/abs/1910.01108}
\BIBentrySTDinterwordspacing

\bibitem{liu2019}
\BIBentryALTinterwordspacing
Y.~Liu, M.~Ott, N.~Goyal, J.~Du, M.~Joshi, D.~Chen, O.~Levy, M.~Lewis,
  L.~Zettlemoyer, and V.~Stoyanov, ``Roberta: {A} robustly optimized {BERT}
  pretraining approach,'' \emph{CoRR}, vol. abs/1907.11692, 2019. [Online].
  Available: \url{http://arxiv.org/abs/1907.11692}
\BIBentrySTDinterwordspacing

\bibitem{clark2020}
\BIBentryALTinterwordspacing
K.~Clark, M.-T. Luong, Q.~V. Le, and C.~D. Manning, ``{ELECTRA}: Pre-training
  text encoders as discriminators rather than generators,'' in \emph{ICLR},
  2020. [Online]. Available: \url{https://openreview.net/pdf?id=r1xMH1BtvB}
\BIBentrySTDinterwordspacing

\bibitem{yang2019}
\BIBentryALTinterwordspacing
Z.~Yang, Z.~Dai, Y.~Yang, J.~Carbonell, R.~R. Salakhutdinov, and Q.~V. Le,
  ``Xlnet: Generalized autoregressive pretraining for language understanding,''
  in \emph{Advances in Neural Information Processing Systems}, H.~Wallach,
  H.~Larochelle, A.~Beygelzimer, F.~d\textquotesingle Alch\'{e}-Buc, E.~Fox,
  and R.~Garnett, Eds., vol.~32.\hskip 1em plus 0.5em minus 0.4em\relax Curran
  Associates, Inc., 2019. [Online]. Available:
  \url{https://proceedings.neurips.cc/paper/2019/file/dc6a7e655d7e5840e66733e9ee67cc69-Paper.pdf}
\BIBentrySTDinterwordspacing

\bibitem{severyn2015}
\BIBentryALTinterwordspacing
A.~Severyn and A.~Moschitti, ``{UNITN}: Training deep convolutional neural
  network for {T}witter sentiment classification,'' in \emph{Proceedings of the
  9th International Workshop on Semantic Evaluation ({S}em{E}val 2015)}.\hskip
  1em plus 0.5em minus 0.4em\relax Denver, Colorado: Association for
  Computational Linguistics, Jun. 2015, pp. 464--469. [Online]. Available:
  \url{https://www.aclweb.org/anthology/S15-2079}
\BIBentrySTDinterwordspacing

\bibitem{SpindeMTL}
T.~Spinde, J.-D. Krieger, T.~Ruas, J.~Mitrović, F.~Götz-Hahn, A.~Aizawa, and
  B.~Gipp, ``Exploiting transformer-based multitask learning for the detection
  of media bias in news articles,'' in \emph{Proceedings of the iConference
  2022 [in review]}, February 2021.

\bibitem{ribeiro2020beyond}
\BIBentryALTinterwordspacing
M.~T. Ribeiro, T.~Wu, C.~Guestrin, and S.~Singh, ``Beyond accuracy: Behavioral
  testing of nlp models with checklist,'' in \emph{Association for
  Computational Linguistics (ACL)}, 2020. [Online]. Available:
  \url{https://arxiv.org/abs/2005.04118}
\BIBentrySTDinterwordspacing

\bibitem{vig2019}
\BIBentryALTinterwordspacing
J.~Vig, ``A multiscale visualization of attention in the transformer model,''
  in \emph{Proceedings of the 57th Annual Meeting of the Association for
  Computational Linguistics: System Demonstrations}.\hskip 1em plus 0.5em minus
  0.4em\relax Florence, Italy: Association for Computational Linguistics, Jul.
  2019, pp. 37--42. [Online]. Available:
  \url{https://www.aclweb.org/anthology/P19-3007}
\BIBentrySTDinterwordspacing

\end{thebibliography}
%

\end{document}